\documentclass[11pt]{article}
\PassOptionsToPackage{hyperfootnotes=false}{hyperref}
\usepackage[preprint]{acl}
\usepackage[T1]{fontenc}
\usepackage{times}
\usepackage{booktabs,amsmath,amssymb,graphicx,array}
\usepackage{float}
\usepackage{fvextra}
\DefineVerbatimEnvironment{Prompt}{Verbatim}{fontsize=\small,fontfamily=ptm,breaklines=true,breakanywhere=true}

\newcommand{\AstraGap}{8.33}
\newcommand{\AstraSPE}{0.830}

\newcommand{\AstraStressGap}{70.83}

\newcommand{\AstraEconomy}{0.830}

\newcommand{\AstraDepthThirtyEconomy}{0.820}
\newcommand{\AstraDepthFortyEconomy}{0.630}

\newcommand{\FableGap}{9.65}
\newcommand{\FableSPE}{0.807}

\newcommand{\FableStressGap}{80.28}

\newcommand{\FableEconomy}{0.807}

\newcommand{\FableDepthThirtyEconomy}{0.790}
\newcommand{\FableDepthFortyEconomy}{0.600}

\newcommand{\OpusGap}{13.01}
\newcommand{\OpusSPE}{0.745}

\newcommand{\OpusStressGap}{99.22}

\newcommand{\OpusEconomy}{0.745}

\newcommand{\OpusDepthThirtyEconomy}{0.730}
\newcommand{\OpusDepthFortyEconomy}{0.540}

\newcommand{\SolGap}{28.44}
\newcommand{\SolSPE}{0.466}

\newcommand{\SolStressCompletion}{64.2}
\newcommand{\SolStressGap}{182.45}

\newcommand{\SolReliability}{98.3}
\newcommand{\SolCoordination}{85.0}
\newcommand{\SolEconomy}{0.558}
\newcommand{\SolDepthTenCoordination}{94.2}
\newcommand{\SolDepthFortyCoordination}{76.6}
\newcommand{\SolDepthTenEconomy}{0.710}

\newcommand{\SolDepthFortyEconomy}{0.350}

\newcommand{\SolBlockZeroCompletion}{87.1}
\newcommand{\SolBlockThreeCompletion}{77.9}

\newcommand{\SonnetGap}{30.52}
\newcommand{\SonnetSPE}{0.456}

\newcommand{\SonnetStressCompletion}{67.1}
\newcommand{\SonnetStressGap}{188.50}

\newcommand{\SonnetReliability}{98.2}
\newcommand{\SonnetCoordination}{85.7}
\newcommand{\SonnetEconomy}{0.541}
\newcommand{\SonnetDepthTenCoordination}{92.5}
\newcommand{\SonnetDepthFortyCoordination}{79.5}
\newcommand{\SonnetDepthTenEconomy}{0.690}

\newcommand{\SonnetDepthFortyEconomy}{0.340}

\newcommand{\LunaSPE}{0.377}

\newcommand{\LunaStressCompletion}{10.0}

\newcommand{\LunaMeanCompleteDepth}{21.9}
\newcommand{\LunaReliability}{95.2}
\newcommand{\LunaCoordination}{63.0}
\newcommand{\LunaEconomy}{0.629}
\newcommand{\LunaDepthTenCoordination}{85.2}
\newcommand{\LunaDepthFortyCoordination}{41.7}

\newcommand{\LunaBlockZeroCompletion}{66.2}
\newcommand{\LunaBlockThreeCompletion}{51.7}

\newcommand{\HaikuSPE}{0.344}

\newcommand{\HaikuStressCompletion}{11.7}

\newcommand{\HaikuMeanCompleteDepth}{21.4}
\newcommand{\HaikuReliability}{96.6}
\newcommand{\HaikuCoordination}{58.0}
\newcommand{\HaikuEconomy}{0.615}
\newcommand{\HaikuDepthTenCoordination}{80.4}
\newcommand{\HaikuDepthFortyCoordination}{31.9}

\title{Verifiable Efficiency Limits of Innate LLM Planning in Sliding Puzzles}
\author{
  Yezhou Cheng \\
  Uber, Inc. \\
  \texttt{yzcheng.ai@gmail.com}
  \And
  Charles Ling\thanks{Corresponding author.} \\
  Western University \\
  \texttt{charles.ling@uwo.ca}
}
\hypersetup{
  pdftitle={Verifiable Efficiency Limits of Innate LLM Planning in Sliding Puzzles},
  pdfauthor={Yezhou Cheng and Charles Ling},
  pdfsubject={Planning evaluation in sliding puzzles},
  pdfkeywords={large language models, planning, sliding puzzles, plan efficiency}
}
\begin{document}
\maketitle

\begin{abstract}
Tool-assisted planning success can reflect external search and
verification, leaving a model's unaided planning ability unresolved.
We isolate innate planning through complete plans generated in one
response without tools or execution feedback. We propose a verifiable
framework that decomposes success-weighted plan efficiency as
$\mathrm{SPE}=R\times Q\times E$: execution reliability, planning
coordination among legal plans, and action economy among completed
plans. Trajectory replay and certified optima verify these factors;
the joint score includes unsuccessful attempts. We evaluate 8,400
full-plan outputs from seven model conditions on $4\times4$ sliding
puzzles, controlling exact horizon and obstacle count in the primary
cohort. Models that complete every trial still become less economical
at longer horizons, while higher efficiency among completed plans can
coexist with poor goal attainment. SPE separates the overall capability
hierarchy, while its factors distinguish coordination failure from
inefficient completion. The framework makes long-horizon limitations
in unaided planning quantitatively verifiable.
\end{abstract}

\section{Introduction}
A long plan must coordinate many actions, preserve their preconditions,
and reach a specified final state. An early mistake can invalidate later
actions; a legal sequence can still stop short or waste effort.
Evaluating only success or only the length of successful solutions
leaves these distinctions unresolved.

PlanBench evaluates valid and cost-optimal plans
\citep{valmeekam2023planbench}. Puzzle evaluations examine accuracy
across complexity \citep{shojaee2025illusion}, while
\citet{schepanowski2025limits} report invalid moves, noncompletion,
and actual versus optimal solution lengths. These established quantities
motivate a more specific question: \emph{where does planning fail when
successful-path efficiency and overall success disagree?}
We propose a verifiable framework that couples a conditional three-level
decomposition with exact-horizon and obstacle controls. It connects
overall performance with the requirement that fails, while retaining
unsuccessful attempts in efficiency comparisons.

We evaluate complete plans for ordinary and blocked sliding puzzles.
The evaluator replays each action and compares completed plans with
a certified optimum. Exact shortest length provides a common horizon
scale. A balanced grid of lengths and obstacle counts separates a longer
minimum plan from a changed movement graph. Increasing the number
of blocks is not assumed to increase difficulty: it also removes tiles
and changes the available routes.

The framework decomposes success-weighted plan efficiency as
$\mathrm{SPE}=R\times Q\times E$: execution reliability, planning
coordination, and action economy. Nested conditioning separates invalid
execution, legal noncompletion, and inefficient completion. SPE compares
overall performance; its factors explain the deficit.
\emph{Innate planning} means one response without external search,
tools, execution feedback, or repair.

\section{Verifiable Full-Plan Evaluation}
\subsection{Tasks and exact horizon control}
A board contains one blank, numbered tiles, and $k\in\{0,1,2,3\}$
fixed blocked cells. An action $U,D,L,$ or $R$ moves the blank by one
cell and swaps it with the adjacent tile. Entering a block or leaving
the board is illegal. The goal lists numbered tiles in ascending
row-major order across available cells, with the blank last. The model
receives the initial and goal boards and returns the entire move sequence.
No intermediate board states or shortest paths are supplied.

The primary cohort contains 320 distinct boards: 20 for each cell of
$k\in\{0,1,2,3\}$ crossed with $d^*\in\{10,20,30,40\}$.
Within each cell, boards are balanced across the selected layouts
(Appendix~\ref{app:boards}). Obstacle count is balanced within each
certified primary horizon; detailed comparisons are reported in
Appendix~\ref{app:extra}. A separate cohort contains 80 boards at
$d^*\in\{50,60\}$ and $k\in\{2,3\}$, again with 20 per cell.
The stress cohort changes obstacle composition; it is not pooled into
the primary horizon comparison.

Each optimum has a checkable certificate. Let $h(s)$ be the sum of
shortest free-cell graph distances from each numbered tile to its goal
cell. Every action moves exactly one tile along an edge, so it can
reduce $h$ by at most one; any solution needs at least $h(s)$ moves.
We retain boards with a legal goal-reaching witness of length $h(s)$.
The lower and upper bounds agree, certifying $d^*=h(s)$ without trusting
the model. A separate implementation verifies the graph distances and
replays the witnesses for all 400 boards.

This construction controls exact horizon but selects a restricted
family: an optimal solution can reduce $h$ at every step. It therefore
tests coordination of long sequences without adding the distinct
challenge of mandatory heuristic retreat. More general boards require
an exact search oracle and are outside this cohort.

\subsection{Three levels of verifiable limitation}
For output $j$, let $L_j$ be its action count and $S_j=1$ if the entire
sequence is legal and ends at the goal. The verifier stops at the first
illegal action without skipping or repairing it; $B_j=1$ denotes such
an error. Completion, legal noncompletion, and illegal plans partition
all outputs. Reaching the goal partway through a sequence is insufficient.
With $N$ outputs, completion and plan-level illegality are
\begin{equation}
C=\frac1N\sum_jS_j,\qquad I=\frac1N\sum_jB_j.
\end{equation}
\paragraph{Execution reliability.}
$R=1-I$ measures the fraction of wholly legal plans. This audits explicit
transition rules, rather than an action-level error probability.
\paragraph{Planning coordination.}
$Q=C/R$ measures goal attainment among legal plans: can executable actions
form a complete goal-directed sequence? Conditioning on legality separates
this behavioral deficit from execution errors. It does not identify an
internal cognitive mechanism. $Q$ is undefined if no plan is legal.
\paragraph{Action economy.}
$E$ is the mean $d_j^*/L_j$ among completed plans. We also report
mean excess moves against the certified optimum:
\begin{equation}
G=\frac{\sum_{j:S_j=1}(L_j-d_j^*)}{\sum_jS_j}.
\end{equation}
$E$ and $G$ are undefined if no plan succeeds. Failed plans have no
optimal gap; assigning them zero would reward failure. $O$ denotes
optimal completions over all outputs. Appendix~\ref{app:extra} retains
completion, optimal-completion, and illegal-plan rates.

To compare overall performance, define success-weighted plan efficiency:
\begin{equation}
\mathrm{SPE}=\frac1N\sum_j
\begin{cases}d_j^*/L_j,&S_j=1,\\0,&S_j=0.\end{cases}
\label{eq:spe}
\end{equation}
When $C>0$, $\mathrm{SPE}=R\times Q\times E=C\times E$ exactly.
The nested denominators cancel: legal plans over all outputs,
completed plans over legal plans, and mean economy among completions.
This identity requires no independence assumption. It satisfies
$0\leq\mathrm{SPE}\leq C\leq1$. Its components show which requirement
limits overall performance.

\subsection{Matched conditions and uncertainty}
We compare Astra, Fable~5, Opus~5.5, GPT-5.6~Sol, Claude~Sonnet~5.5, GPT-5.6~Luna, and Claude~Haiku~4.5 on identical
boards, with three trials per board. Each condition has 960 primary
outputs and 240 stress outputs, for 8,400 total. The protocol requests
the shortest complete plan in a fresh context, with no tools or feedback;
Appendix~\ref{app:protocol} gives the prompt specification.
Figures and tables use the family name and version.
All substantive failures are retained.

We report 95\% percentile bootstrap intervals from 5,000 draws.
Whole boards are resampled within each block-count--depth cell,
keeping their three trials together and preserving the balanced design.
The same sampled board indices are used across conditions. These
intervals characterize variation within the sampled design; zero
observed errors do not certify zero population error probability.

\begin{table*}[t]
\centering
\setlength{\tabcolsep}{8pt}
\begin{tabular}{@{}lrrrrr@{}}
\toprule
Condition & Execution $R$ $\uparrow$ & Coordination $Q$ $\uparrow$ & Economy $E$ $\uparrow$ & Gap $G$ $\downarrow$ & Joint SPE $\uparrow$\\
 & (\%) & (\%) & ($0$--$1$) & (moves) & ($0$--$1$) \\
\midrule
Astra & 100.0 & 100.0 & 0.830 & 8.33 & 0.830 \\
Fable 5 & 100.0 & 100.0 & 0.807 & 9.65 & 0.807 \\
Opus 5.5 & 100.0 & 100.0 & 0.745 & 13.01 & 0.745 \\
Sol 5.6 & 98.3 & 85.0 & 0.558 & 28.44 & 0.466 \\
Sonnet 5.5 & 98.2 & 85.7 & 0.541 & 30.52 & 0.456 \\
Luna 5.6 & 95.2 & 63.0 & 0.629 & 19.13 & 0.377 \\
Haiku 4.5 & 96.6 & 58.0 & 0.615 & 19.80 & 0.344 \\
\bottomrule
\end{tabular}
\caption{Primary results pooled over $d^*\in\{10,20,30,40\}$
(320 boards, three trials per condition). $R$ uses all 960 outputs;
$Q$ uses legal outputs;
$E$ and $G$ use completed outputs. Joint SPE equals $RQE$ before rounding and includes
every failure. Higher economy among completed plans can coexist with
lower joint SPE.}
\label{tab:primary}

\end{table*}

\begin{figure*}[t]
\centering
\includegraphics[width=\textwidth]{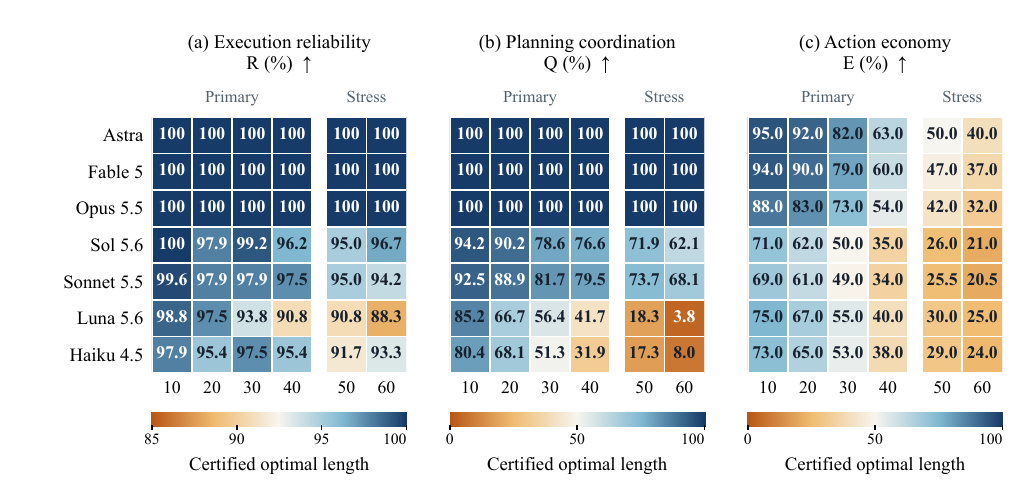}
\caption{Three factors of $\mathrm{SPE}=R\times Q\times E$ across
certified horizons, with one row per model. All factors are shown as
percentages to one decimal; $E$ is mean $d^*/L$ among completions.
Blue indicates better values and orange poorer values on each panel's
stated linear scale. The white gutter separates primary cohorts
(240 outputs per condition and length) from stress cohorts
(120 outputs, only two or three blocks). $Q$ conditions on legal plans.}
\label{fig:horizon}
\end{figure*}

\section{Results}
\subsection{Execution, coordination, and economy separate limits}
Table~\ref{tab:primary} separates the three requirements. Astra,
Fable~5, and Opus~5.5 have $R=Q=1$: every output is legal and reaches
the goal. Their economy is \AstraEconomy{}, \FableEconomy{}, and
\OpusEconomy{}, with primary gaps of \AstraGap{}, \FableGap{}, and
\OpusGap{} moves. Identical execution and coordination therefore
conceal differences in action economy.

Sol~5.6 and Sonnet~5.5 retain execution reliability of
\SolReliability\% and \SonnetReliability\%, but only
\SolCoordination\% and \SonnetCoordination\% of their legal plans
reach the goal. Their economy is \SolEconomy{} and \SonnetEconomy{},
with gaps of \SolGap{} and \SonnetGap{} moves. Both goal
coordination and economical completion limit them.

Luna~5.6 and Haiku~4.5 have reliability of \LunaReliability\% and
\HaikuReliability\%, but coordination of only \LunaCoordination\%
and \HaikuCoordination\%. Economy among completions is
\LunaEconomy{} and \HaikuEconomy{}. Legal noncompletion exposes
a further limitation after execution errors have been excluded.

\subsection{Longer horizons expose coordination and economy limits}
Figure~\ref{fig:horizon} compares $R$, $Q$, and $E$ while keeping
primary obstacle proportions fixed. Sol~5.6's coordination falls from
\SolDepthTenCoordination\% at $d^*=10$ to
\SolDepthFortyCoordination\% at $d^*=40$; Sonnet~5.5's falls from
\SonnetDepthTenCoordination\% to \SonnetDepthFortyCoordination\%.
Their economy also falls: \SolDepthTenEconomy{} to
\SolDepthFortyEconomy{} for Sol~5.6, and \SonnetDepthTenEconomy{}
to \SonnetDepthFortyEconomy{} for Sonnet~5.5.

Luna~5.6's coordination falls from \LunaDepthTenCoordination\% to
\LunaDepthFortyCoordination\%; Haiku~4.5's falls from
\HaikuDepthTenCoordination\% to \HaikuDepthFortyCoordination\%.
At $d^*=40$, most legal attempts therefore fail to reach the goal.
Their economy also declines, but remains above Sol~5.6 and Sonnet~5.5
at every primary depth. This conditional advantage describes
completed attempts; it does not erase legal noncompletion.

Astra, Fable~5, and Opus~5.5 retain $R=Q=1$ at every tested length.
Their $E$ falls from \AstraDepthThirtyEconomy{},
\FableDepthThirtyEconomy{}, and \OpusDepthThirtyEconomy{} at length
30 to \AstraDepthFortyEconomy{}, \FableDepthFortyEconomy{}, and
\OpusDepthFortyEconomy{} at length 40. Opus~5.5 has lower economy
than Astra and Fable~5 at every primary depth. Perfect completion
thus coexists with a measurable decline in relative action efficiency.

For an illustrative whole-plan reliability target, let $H_{.9}$ be the
largest tested primary length with $C\geq90\%$ at every tested length
up to it. Astra, Fable~5, and Opus~5.5 qualify through 40; Sol~5.6
and Sonnet~5.5 through 10. Luna~5.6 and Haiku~4.5 do not qualify
at the shortest tested length. These finite-grid summaries do not imply
impossibility on longer individual boards.

\subsection{Joint scores reveal the overall capability hierarchy}
Primary joint scores are \AstraSPE{}, \FableSPE{}, and \OpusSPE{}
for the stronger conditions; \SolSPE{} and \SonnetSPE{} for the
intermediate conditions; and \LunaSPE{} and \HaikuSPE{} for the
lower-tier conditions. The product of execution reliability, planning
coordination, and action economy makes the overall hierarchy explicit.

Luna~5.6 and Haiku~4.5 have higher conditional economy
(\LunaEconomy{} and \HaikuEconomy{}) than Sol~5.6 and Sonnet~5.5
(\SolEconomy{} and \SonnetEconomy{}), but remain below the stronger
conditions. Their coordination deficits outweigh this conditional
advantage, leaving their joint scores lower. Overall capability and
individual conditional metrics need not have identical rankings.

Success-only summaries also select different boards. The primary mean
depth is 25; completed Luna~5.6 and Haiku~4.5 outputs average
\LunaMeanCompleteDepth{} and \HaikuMeanCompleteDepth{}.
Shorter boards are overrepresented among their successes. Their
depth-specific economy advantage over the intermediate conditions
shows that this selection is not the sole source of the conditional ranking.

The separate blocked stress cohort reinforces this distinction. Astra,
Fable~5, and Opus~5.5 again have perfect completion, with gaps of
\AstraStressGap{}, \FableStressGap{}, and \OpusStressGap{} moves.
Sol~5.6 and Sonnet~5.5 complete \SolStressCompletion\% and
\SonnetStressCompletion\%, with gaps of \SolStressGap{} and
\SonnetStressGap{}. Luna~5.6 and Haiku~4.5 complete only
\LunaStressCompletion\% and \HaikuStressCompletion\%. Their economy
remains above the intermediate conditions, but their joint scores
remain lower. Changed obstacle composition prevents attributing the
primary-to-stress difference solely to horizon.

\section{Discussion and Conclusion}
Assessing innate planning requires separating a model's unaided
performance from external assistance. Our single-response protocol
makes this separation explicit, and our verifiable framework quantifies
the resulting limitations through $\mathrm{SPE}=R\times Q\times E$.
Trajectory replay verifies execution reliability and goal coordination;
certified optima verify action economy. SPE measures overall
performance, while its factors identify which requirement limits it.

The results show why this joint diagnosis matters. Astra, Fable~5,
and Opus~5.5 complete every plan yet lose action economy at longer
horizons. Sol~5.6 and Sonnet~5.5 lose both coordination and economy.
Luna~5.6 and Haiku~4.5 attain slightly higher economy than these
intermediate conditions, but greater coordination deficits yield lower
SPE. Completion and action efficiency thus expose distinct planning
limits. The contribution is the unified diagnostic framework and its
controlled, verifiable evaluation design.

Future work will extend the framework to planning games such as
Sokoban, using domain-specific transition verifiers and certified
optima to test whether the same diagnoses transfer beyond sliding
puzzles. Paired evaluations of unaided and tool-assisted planning on
identical instances can then identify which factors external support
improves. These extensions connect the limits of unaided planning to
the mechanisms that help overcome them.

\clearpage
\section*{Limitations}
\paragraph{Deployment costs.}
Our framework evaluates planning performance in action space.
Token consumption, response latency, and monetary cost could complement
this evaluation when deployment efficiency is of interest.

\paragraph{Trajectory detail.}
Additional trajectory diagnostics, such as first-error positions and
repeated-state visits, could provide finer descriptions of observed
behavior. These extensions would enrich the analysis while preserving
the definitions and exact decomposition of SPE.

\paragraph{Model coverage.}
The current evaluation covers a selected set of model families.
Extending the comparison to additional families, including open-weight
models, would broaden the empirical coverage and support comparisons
across model scales and deployment settings. These models can be
evaluated using the same verification protocol and SPE decomposition.

\clearpage
\appendix
\raggedbottom
\section{Board Construction and Certificates}
\label{app:boards}
Blocked positions below use zero-based row-major indexing.
The primary cohort has 20 boards at each block-count--depth cell:
\begin{itemize}
\item $k=0$: no blocks; 20 boards per depth.
\item $k=1$: positions $\{5\},\{6\},\{9\},\{10\}$; five boards
per layout and depth.
\item $k=2$: positions $\{5,10\},\{6,9\}$; ten boards per layout
and depth.
\item $k=3$: four layouts, with five boards per layout and depth.
The blocked sets are $\{5,6,9\}$, $\{5,6,10\}$,
$\{5,9,10\}$, and $\{6,9,10\}$.
\end{itemize}
The stress cohort uses the same $k=2,3$ layouts and quotas at
depths 50 and 60. All layouts have a connected free-cell graph.
Goals have $15-k$ numbered tiles, one blank, and the specified blocks.

Candidate generation starts at the goal and randomly chooses legal
actions that increase the tile graph-distance sum by one. Proposals
that stop before the target length or duplicate an accepted board
are discarded. Reversing a length-$d$ accepted walk yields a solution
of length $d$, while the initial lower bound is also $d$.
Selection uses seed 20261010. The retained bank stores initial states,
goals, block positions, proposal indices, graph-distance bounds, and
complete certificate paths. The paths are evaluation data and are
excluded from the planning input.

\paragraph{Proof of the lower bound.}
Let $V$ be the free cells and $\delta_V$ their shortest-path distance.
For numbered tile $t$, let $p_t$ be its current cell and $q_t$ its
goal cell. Set $h(s)=\sum_t\delta_V(p_t,q_t)$, excluding the blank.
In a legal move only one tile changes position, across one graph
edge. By the triangle inequality its distance to $q_t$ decreases
by at most one. Since the goal has $h=0$, any solution takes at
least $h(s)$ actions. A replayed solution of length $h(s)$ attains
this bound and is therefore shortest.

\section{Protocol and Verification Specification}
\label{app:protocol}
Each trial requests a complete shortest plan from the initial state
to the explicit goal. Contexts are independent across boards and
trials. No tools, demonstrations, execution feedback, or oracle values
are included. The response format consists of a single complete move
list; the direction convention always refers to the blank.

\begin{Prompt}
You are solving a 4-by-4 sliding puzzle.
0 is the blank. -1 is a fixed blocked cell.
U/D/L/R move the blank one cell up/down/left/right
and swap it with the adjacent numbered tile.
Never move outside the board or into a blocked cell.
Find a shortest complete sequence reaching the goal.
Return only MOVES: followed by comma-separated U,D,L,R.

Initial board:
[four rows of integers]
Goal board:
[four rows of integers]
\end{Prompt}

The verifier reads the explicit move string, checks each target cell,
and stops at the first illegal action. Legal prefixes and first-error
positions remain available for supplementary audits. A second
implementation independently verifies 400 optimum certificates and
replays all 8,400 outputs, reproducing every score. The data retain
complete move strings rather than only outcome labels.

\clearpage
\section{Supplementary Results}
\label{app:extra}
\begin{figure}[H]
\centering
\includegraphics[width=\columnwidth]{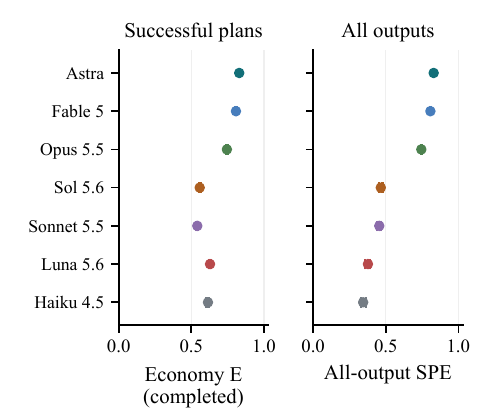}
\caption{Conditional $E$ and joint SPE on primary boards. Luna~5.6
and Haiku~4.5 have higher $E$ but lower SPE than Sol~5.6 and
Sonnet~5.5. SPE includes all failures. Bars show 95\% intervals from
stratified whole-board bootstrap resampling, keeping the three trials together.}
\label{fig:joint}
\end{figure}

Tables~\ref{tab:counts} and~\ref{tab:raw} give primary outcome
counts and rates. Table~\ref{tab:blocks} reports completion by block count,
averaging equally over the four primary depths. Tables~\ref{tab:stress}
and~\ref{tab:intervals} provide stress results and primary uncertainty.
Figure~\ref{fig:joint} contrasts conditional economy with joint performance.

\begin{table}[H]
\centering\small
\setlength{\tabcolsep}{4pt}
\begin{tabular}{@{}lrrr@{}}
\toprule
Condition & Complete & Legal non-goal & Illegal\\
\midrule
Astra & 960 & 0 & 0 \\
Fable 5 & 960 & 0 & 0 \\
Opus 5.5 & 960 & 0 & 0 \\
Sol 5.6 & 802 & 142 & 16 \\
Sonnet 5.5 & 808 & 135 & 17 \\
Luna 5.6 & 576 & 338 & 46 \\
Haiku 4.5 & 538 & 389 & 33 \\
\bottomrule
\end{tabular}
\caption{Mutually exclusive primary outcomes, summing to 960 per
condition. The illegal column includes both boundary and obstacle
violations.}
\label{tab:counts}
\end{table}

\begin{table}[H]
\centering\small
\setlength{\tabcolsep}{5pt}
\begin{tabular}{@{}lrrr@{}}
\toprule
Condition & $C$ (\%) & $O$ (\%) & $I$ (\%)\\
\midrule
Astra & 100.0 & 26.7 & 0.0 \\
Fable 5 & 100.0 & 24.0 & 0.0 \\
Opus 5.5 & 100.0 & 10.1 & 0.0 \\
Sol 5.6 & 83.5 & 0.0 & 1.7 \\
Sonnet 5.5 & 84.2 & 0.0 & 1.8 \\
Luna 5.6 & 60.0 & 0.0 & 4.8 \\
Haiku 4.5 & 56.0 & 0.8 & 3.4 \\
\bottomrule
\end{tabular}
\caption{Primary completion, optimal completion, and illegal-plan
percentages. All three use 960 outputs per condition.}
\label{tab:raw}
\end{table}

\newpage
\paragraph{Obstacle-count comparisons.}
At each primary depth, all four block counts are represented.
Table~\ref{tab:blocks} shows that Luna~5.6's completion is
\LunaBlockZeroCompletion\% with no blocks and
\LunaBlockThreeCompletion\% with three; Sol~5.6's is
\SolBlockZeroCompletion\% and \SolBlockThreeCompletion\%.
Other contrasts are smaller and not uniformly monotonic. Blocks jointly
change tile count and movement topology, so their count alone does not
define difficulty.

\begin{table}[H]
\centering\small
\setlength{\tabcolsep}{6pt}
\begin{tabular}{@{}lrrrr@{}}
\toprule
Condition & $k=0$ & $k=1$ & $k=2$ & $k=3$\\
\midrule
Astra & 100.0 & 100.0 & 100.0 & 100.0 \\
Fable 5 & 100.0 & 100.0 & 100.0 & 100.0 \\
Opus 5.5 & 100.0 & 100.0 & 100.0 & 100.0 \\
Sol 5.6 & 87.1 & 86.2 & 82.9 & 77.9 \\
Sonnet 5.5 & 84.2 & 85.4 & 83.8 & 83.3 \\
Luna 5.6 & 66.2 & 65.0 & 57.1 & 51.7 \\
Haiku 4.5 & 63.7 & 56.2 & 51.7 & 52.5 \\
\bottomrule
\end{tabular}
\caption{Primary completion percentages by block count. Each
condition has 240 outputs at each count, balanced across depths.}
\label{tab:blocks}
\end{table}

\begin{table}[H]
\centering\small
\setlength{\tabcolsep}{4.5pt}
\begin{tabular}{@{}lrrrr@{}}
\toprule
Condition & $C$ (\%) & $I$ (\%) & $G$ (moves) & SPE\\
\midrule
Astra & 100.0 & 0.0 & 70.83 & 0.450 \\
Fable 5 & 100.0 & 0.0 & 80.28 & 0.420 \\
Opus 5.5 & 100.0 & 0.0 & 99.22 & 0.370 \\
Sol 5.6 & 64.2 & 4.2 & 182.45 & 0.152 \\
Sonnet 5.5 & 67.1 & 5.4 & 188.50 & 0.155 \\
Luna 5.6 & 10.0 & 10.4 & 129.58 & 0.029 \\
Haiku 4.5 & 11.7 & 7.5 & 144.57 & 0.032 \\
\bottomrule
\end{tabular}
\caption{Separate stress cohort: 80 boards, 240 outputs per
condition, balanced across $k=2,3$ and $d^*=50,60$. Gap remains
conditional on successful completion.}
\label{tab:stress}
\end{table}

\begin{table}[H]
\centering\small
\setlength{\tabcolsep}{4pt}
\begin{tabular}{@{}lrr@{}}
\toprule
Condition & $C$ interval (\%) & $G$ interval (moves)\\
\midrule
Astra & $[100.0,100.0]$ & $[8.12,8.54]$ \\
Fable 5 & $[100.0,100.0]$ & $[9.43,9.88]$ \\
Opus 5.5 & $[100.0,100.0]$ & $[12.73,13.29]$ \\
Sol 5.6 & $[81.2,85.7]$ & $[27.49,29.35]$ \\
Sonnet 5.5 & $[81.8,86.5]$ & $[29.58,31.44]$ \\
Luna 5.6 & $[57.0,62.9]$ & $[17.90,20.31]$ \\
Haiku 4.5 & $[53.1,58.9]$ & $[18.52,21.13]$ \\
\bottomrule
\end{tabular}
\caption{Primary 95\% percentile intervals from 5,000 stratified
whole-board bootstrap draws. Random seed is 20261012. All three
trials of each resampled board are kept together.}
\label{tab:intervals}
\end{table}

\newpage
\begin{thebibliography}{3}
\providecommand{\natexlab}[1]{#1}

\bibitem[{Schepanowski and Ling(2025)}]{schepanowski2025limits}
Charles Schepanowski and Charles Ling. 2025.
\newblock \href {https://arxiv.org/abs/2511.21591} {On the limits of innate
  planning in large language models}.
\newblock \emph{Preprint}, arXiv:2511.21591.

\bibitem[{Shojaee et~al.(2025)Shojaee, Mirzadeh, Alizadeh, Horton, Bengio, and
  Farajtabar}]{shojaee2025illusion}
Parshin Shojaee, Iman Mirzadeh, Keivan Alizadeh, Maxwell Horton, Samy Bengio,
  and Mehrdad Farajtabar. 2025.
\newblock \href {https://arxiv.org/abs/2506.06941} {The illusion of thinking:
  Understanding the strengths and limitations of reasoning models via the lens
  of problem complexity}.
\newblock \emph{Advances in Neural Information Processing Systems}.

\bibitem[{Valmeekam et~al.(2023)Valmeekam, Marquez, Olmo, Sreedharan, and
  Kambhampati}]{valmeekam2023planbench}
Karthik Valmeekam, Matthew Marquez, Alberto Olmo, Sarath Sreedharan, and
  Subbarao Kambhampati. 2023.
\newblock \href {https://arxiv.org/abs/2206.10498} {Planbench: An extensible
  benchmark for evaluating large language models on planning and reasoning
  about change}.
\newblock \emph{Advances in Neural Information Processing Systems}.

\end{thebibliography}
\end{document}